\documentclass[11pt]{article}
\usepackage[utf8]{inputenc}
\usepackage[T1]{fontenc}
\usepackage{lmodern}
\usepackage{graphicx}
\usepackage[style=base,figurename=Fig.,labelfont=bf,labelsep=period]{caption}
\usepackage{subcaption}
\usepackage{amsmath}
\usepackage{newtxtext,newtxmath}
\usepackage[colorlinks=true,citecolor=red,linkcolor=black]{hyperref}
\usepackage{tikz}
\usetikzlibrary{shapes.geometric}
\usepackage[numbers]{natbib}
\usepackage{geometry}
\usepackage{booktabs}
\usepackage{array}
\usepackage{tabularx}
\usepackage{siunitx}
\title{Interpretable Locomotion Prediction in Construction Using a Memory-Driven LLM Agent With Chain-of-Thought Reasoning}
\author{
  Ehsan Ahmadi\thanks{Ph.D. Candidate, Bert S. Turner Department of Construction Management, Louisiana State University, Baton Rouge, LA 70803. Email: eahmad2@lsu.edu} \\
  Chao Wang\thanks{Associate Professor, Bert S. Turner Department of Construction Management, Louisiana State University, Baton Rouge, LA 70803}
}
\date{\today}

\begin{document}

\maketitle

\begin{abstract}
Construction tasks are inherently unpredictable, with dynamic environments and safety-critical demands posing significant risks to workers. Exoskeletons offer potential assistance but falter without accurate intent recognition across diverse locomotion modes. This paper presents a locomotion prediction agent leveraging Large Language Models (LLMs) augmented with memory systems, aimed at improving exoskeleton assistance in such settings. Using multimodal inputs—spoken commands and visual data from smart glasses—the agent integrates a Perception Module, Short-Term Memory (STM), Long-Term Memory (LTM), and Refinement Module to predict locomotion modes effectively. Evaluation reveals a baseline weighted F1-score of 0.73 without memory, rising to 0.81 with STM, and reaching 0.90 with both STM and LTM, excelling with vague and safety-critical commands. Calibration metrics, including a Brier Score drop from 0.244 to 0.090 and ECE from 0.222 to 0.044, affirm improved reliability. This framework supports safer, high-level human-exoskeleton collaboration, with promise for adaptive assistive systems in dynamic industries.
\end{abstract}

\section{Introduction}
Construction workers face significant risks of work-related musculoskeletal disorders (WMSDs), driven by repetitive tasks, heavy load handling, and non-neutral postures in dynamic, unpredictable environments \cite{al2024integrating,kim2019potential}. In the U.S., construction workers experience an 11\% higher WMSD rate than the average across industries, with the back and shoulders most affected~\cite{kim2019potential}. While exoskeletons show promise in reducing physical strain---passive designs lowering back muscle activity by 10-40\% and active ones achieving up to 80\% reductions across multiple regions~\cite{de2016exoskeletons}---their practical deployment remains limited by discomfort and poor alignment with human movements, particularly in construction settings~\cite{flor2023exoskeletons}. Central to these limitations is the challenge of accurately recognizing user intent across varied tasks, a gap that restricts effective collaboration~\cite{baud2021review,tucker2015control}. This misalignment heightens safety risks, as powered exoskeletons may generate destructive forces if their controlled output deviates from the user’s intent~\cite{tucker2015control}.

Addressing this locomotion intent recognition challenge is pivotal to unlocking effective exoskeleton assistance in construction, particularly for diverse, safety-critical tasks like ladder climbing and obstacle navigation. Traditional evaluation of assistive technologies like lower-limb exoskeletons has focused narrowly on routine tasks such as straight walking~\cite{pinto2020performance}, neglecting these critical locomotion modes and requiring a shift beyond conventional control paradigms that lack flexibility for dynamic contexts. Construction tasks are highly variable, requiring workers to adapt to shifting demands, irregular workflows, and unstructured environments where movement patterns are unpredictable~\cite{kim2019potential}. This variability complicates the implementation of assistive technologies, as rigid control approaches struggle to accommodate rapid task transitions and environmental uncertainty.

To overcome the limitations of traditional control approaches in the dynamic and unpredictable environments of construction, this paper introduces a locomotion mode prediction agent powered by Large Language Models (LLMs), utilizing spoken commands and visual data from smart glasses. LLMs provide a robust foundation, having achieved significant progress in multimodal understanding~\cite{openai2023gpt4v,team2024gemini}. This multimodal perception is critical for accurately interpreting user intent across diverse tasks encountered in construction. Furthermore, LLMs exhibit strong reasoning and planning abilities, facilitated by techniques such as chain-of-thought (CoT) prompting, which enables them to break down complex problems and predict intents step by step~\cite{wei2022chain,li2024chain}. Their pre-training on large-scale corpora allows for few-shot and zero-shot generalization, enabling adaptation to diverse tasks without extensive retraining~\cite{brown2020language,kojima2022large,wei2021finetuned}, a crucial advantage in the constantly evolving demands of construction. By leveraging these strengths, LLM-based agents are capable of natural language interaction, environmental comprehension, generalization, reasoning, planning, and tool usage to perform a wide variety of tasks effectively~\cite{xi2025rise}. Additionally, the integration of a memory module, inspired by human cognitive processes, enhances the agent’s ability to store and retrieve contextual information for more consistent and effective performance~\cite{wang2024survey}.

While LLMs offer significant advantages for various tasks including intent recognition, their limitations pose challenges for safety-critical applications like construction, where reliable decision-making is essential for effective locomotion prediction. LLMs can exhibit undesirable behaviors, such as generating nonfactual information (hallucinations), providing false rationalizations (unfaithful reasoning), or pursuing misaligned objectives, potentially leading to a loss of human control \cite{bang2023multitask,turpin2023language,park2024ai}. In construction, where exoskeleton-assisted workers perform hazardous activities, such errors could result in incorrect locomotion predictions, triggering unsafe movements and risking accidents or injuries. Additionally, LLMs operate as opaque "black-box" systems \cite{zhao2024explainability}, which complicates error diagnosis and accountability. This is critical in settings where understanding decision rationales ensures safety and trust. Thus, implementing LLM-based agents in construction demands thorough validation and transparent design to mitigate these risks and ensure dependable human-exoskeleton collaboration.

To enhance the reliability and transparency of the locomotion prediction agent, we augment the LLM with short-term and long-term memory systems to tackle the dynamic and safety-critical demands of construction tasks. Short-term memory captures recent events, providing immediate context for rapid transitions, while long-term memory retains historical patterns to refine predictions, particularly in safety-sensitive scenarios. The agent incorporates a Perception Module with chain-of-thought reasoning to systematically analyze multimodal inputs—spoken commands and visual data. A Refinement Module enhances accuracy by reprocessing ambiguous predictions using memory-derived insights, ensuring robust outcomes despite vague or discrepant user inputs and reducing the risk of errors. Through thorough testing, this paper evaluates these enhancements, aiming to deliver dependable and interpretable locomotion mode predictions tailored to the safety needs of exoskeleton-assisted construction workflows.

\section{Related Works}

\subsection{Locomotion Prediction in Assistive Technologies}
Effective control of assistive technologies, such as exoskeletons, hinges on the ability to predict locomotion accurately, enabling these devices to adapt seamlessly to users’ movements in real time. This involves identifying diverse locomotion modes—like walking, stair climbing, ramp navigation, sitting, and standing—as well as estimating gait phases and handling transitions between movements \cite{novak2015survey, young2015classification, qian2022predictive, yu2023artificial}. Past research has explored a range of techniques, from conventional machine learning and neural network models to more integrated multimodal systems, each leveraging various sensor inputs to enhance prediction performance.

Early efforts in locomotion prediction often utilized traditional machine learning alongside neural network advancements, relying on sensors such as IMUs, pressure insoles, and encoders. For instance, Long et al. \cite{long2016pso} applied an SVM with Particle Swarm Optimization to classify walking, stair ascent and descent, ramp navigation, and transitions using foot pressure and Attitude and Heading Reference System data. Parri et al. \cite{parri2017real} combined hip joint encoders and pressure insoles with time-based decoding and fuzzy logic to predict sitting, standing, and walking, focusing on mode transitions. Similarly, Kim et al. \cite{kim2017kinematic} used a decision tree with IMUs and load cells to recognize walking, stairs, and ramps, emphasizing foot inclination, while Liu et al. \cite{liu2020real} developed an SVM-based system for real-time torque prediction across walking and standing tasks. Utilizing neural networks, Zhu et al. \cite{zhu2020novel} employed IMUs data to classify walking, stairs, and ramps, and Laschowski et al. \cite{laschowski2022environment} used deep convolutional neural networks (CNNs) with camera data to identify indoor/outdoor walking, stairs, and transitions. Guo et al. \cite{guo2024speech} introduced a GA-CNN model incorporating speech commands for mode recognition and gait phase estimation across sitting, standing, and walking, while Martínez-Pascual et al. \cite{martinez2024gait} applied 1D-CNNs to joint angle trajectories for walking, ramp, and stair classification.

More advanced multimodal approaches have since emerged to enhance prediction robustness by integrating diverse sensor data. Li et al. \cite{li2022fusion} fused eye-tracker and depth camera inputs with deep learning and dynamic time warping to predict walking, stair, and ramp navigation. Qian et al. \cite{qian2022predictive} fused RGBD camera and IMU data for locomotion mode recognition, gait phase estimation, and transitions. Sharma et al. \cite{sharma2022improving} and Tsepa et al. \cite{tsepa2023continuous} utilized LSTMs and KIFNet, respectively, with IMUs and smart glasses data to predict joint angles kinematics during walking across varied environments. Wang et al. \cite{wang2022integral} adopted a GA-CNN model with IMUs and force sensors to recognize walking, standing, sitting, and squatting in real time, optimized via Bayesian methods. Li et al. \cite{li2024multimodal} developed a deep belief network integrating accelerometer, gyroscope, and sEMG data with multiple fusion strategies to predict steady-state locomotion and transitions.

While these approaches demonstrate notable success, they predominantly address structured and predictable environments, often relying on supervised learning tailored to specific tasks. Such methods may fall short in capturing the broad spectrum of activities prevalent in construction settings—such as ladder climbing, obstacle navigation, and low-space movement—where unpredictability and variability dominate. Moreover, prior work has largely overlooked the integration of speech and vision, key modalities through which humans naturally convey intent and interpret their surroundings. By fusing these intuitive inputs, our approach seeks to bridge this gap, offering a more comprehensive and adaptable framework for locomotion mode prediction in complex scenarios like construction.

\subsection{LLM Agents with Memory Integration}
Given the need for multimodal interaction with speech and vision in locomotion prediction, LLM agents with memory provide a promising solution, leveraging past experiences and task transitions to boost adaptability. Recent efforts to integrate memory into these agents have enhanced their ability to retain context, adapt to dynamic tasks, and mimic human behavior via LLMs’ capabilities. Researchers have explored diverse memory approaches, improving reasoning, planning, and interaction across robotics, social simulations, gaming, conversational systems, and recommendation tasks, highlighting the versatility of memory-augmented LLMs.

Memory grounds LLMs in interactive environments in several studies. Rana et al. \cite{rana2023sayplan} present SayPlan, using 3D Scene Graphs (3DSGs) with a memory list of explored nodes to enable scalable robotic task planning, efficiently navigating large-scale settings like multi-floor buildings via semantic searches and iterative replanning with simulator feedback. Park et al. \cite{park2023generative} develop generative agents simulating human-like behavior in a sandbox, employing a long-term memory stream of natural language experiences and cached reflections to drive emergent social behaviors like party planning. Lin et al. \cite{lin2023agentsims} introduce AgentSims, an open-source sandbox where LLM agents store daily experiences as embeddings in a vector database, ensuring consistent socio-economic simulation behavior by recalling past interactions.

Memory also boosts task-oriented coordination and decision-making. Li et al. \cite{li2023metaagents} propose MetaAgents, simulating job fair interactions with a hybrid memory of biographies and experiences as summaries and key terms, supporting reasoning for consistent workflow design. Huang et al. \cite{huang2023memory} offer Memory Sandbox, an interactive system letting users manage conversational memory as manipulable objects, toggling visibility to enhance dialogue coherence. Shinn et al. \cite{shinn2023reflexion} develop Reflexion, where agents store verbal self-reflections in an episodic memory buffer integrating short-term trajectories and long-term lessons, improving coding and decision-making performance.


Other works extend LLM context and personalization via memory hierarchies. Packer et al. \cite{packer2023memgpt} propose MemGPT, an OS-inspired system paging data between a fixed-context window and external storage, enabling extended conversations and document analysis with unbounded context access. Wang et al. \cite{wang2023recmind} develop RecMind, a recommendation agent integrating personalized memory (user history) and world knowledge (item data) via tools, using a self-inspiring algorithm for zero-shot recommendations. Zhong et al. \cite{zhong2024memorybank} introduce MemoryBank, enhancing LLMs with a long-term memory repository of conversation logs, event summaries, and user portraits, powering SiliconFriend for empathetic, personalized dialogues tuned with psychological data.

These advancements underscore memory’s critical role in enhancing LLM agents’ reasoning, planning, and contextual awareness. However, most efforts center on simulation-driven context, with limited exploration of multimodal user interactions, such as speech and vision, in physical environments—particularly for understanding human locomotion. Inspired by these developments in memory-enabled agents, our work integrates short-term and long-term memory with multimodal perception to predict locomotion modes in construction settings, addressing the need for robust, context-aware agents in dynamic, safety-critical environments.

\section{Proposed Methodology}
\subsection{System Architecture}
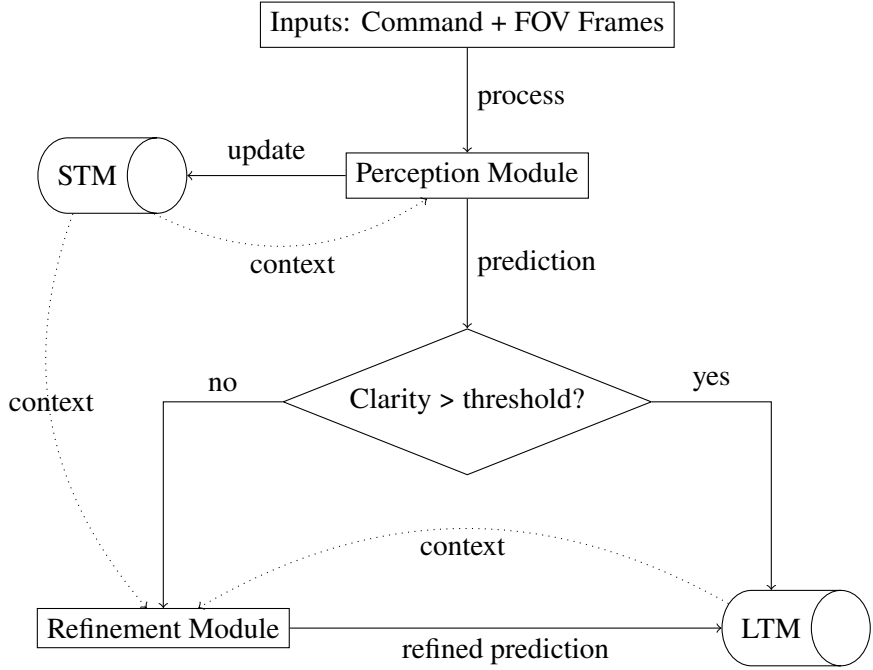
\begin{figure}
\centering
\begin{tikzpicture}
  \node (inputs) at (0,0) [rectangle, draw] {Inputs: Command + FOV Frames};
  \node (perception) at (0,-2) [rectangle, draw] {Perception Module};
  \node (stm) at (-5,-2) [cylinder, draw, aspect=1.5, minimum height=1.5cm, minimum width=1cm] {STM};
  \node (decision) at (0,-5) [diamond, draw, aspect=2.5] {Clarity > threshold?};
  \node (refinement) at (-4,-8) [rectangle, draw] {Refinement Module};
  \node (ltm) at (4,-8) [cylinder, draw, aspect=1.5, minimum height=1.5cm, minimum width=1cm] {LTM};

  \draw [->] (inputs) -- (perception) node[midway, right] {process}; 
  \draw [dotted, ->] (stm) to [bend right=30] node[midway, below] {context} (perception); 
  \draw [->] (perception) -- (decision) node[midway, right] {prediction}; 
  \draw [->] (decision) -| node[pos=0.25, above] {yes} (ltm); 
  \draw [->] (decision) -| node[pos=0.25, above] {no} (refinement); 
  \draw [->] (refinement) -- (ltm) node[midway, below] {refined prediction}; 
  \draw [dotted, ->] (ltm) to [bend right=30] node[midway, below] {context} (refinement); 
  \draw [->] (perception) -- (stm) node[midway, above] {update}; 
  \draw [dotted, ->] (stm) to [bend right=30] node[midway, above] {context} (refinement); 
\end{tikzpicture}
\caption{High-Level Overview of the Agent’s Workflow}
\label{fig:architecture}
\end{figure}

The locomotion prediction agent is designed to interpret multimodal inputs—spoken commands and visual data from smart glasses—to predict a user’s locomotion mode in construction-related activities. It combines a Large Language Model (LLM) with memory systems to address challenges such as dynamic settings, vague or incorrect commands, and safety-critical transitions. The system comprises four core components: the Perception Module, Short-Term Memory (STM), Long-Term Memory (LTM), and the Refinement Module, which collaborate to deliver accurate and safe predictions, as illustrated in Figure \ref{fig:architecture}.

The Perception Module acts as the entry point, processing spoken commands and field-of-view (FOV) frames to generate an initial locomotion mode prediction. Additionally, it evaluates input clarity using metrics like vagueness and discrepancy. Short-Term Memory (STM) maintains a time-sensitive record of recent events to provide immediate context for ongoing tasks and safety-critical transitions. Long-Term Memory (LTM) stores events as vector-based entries and provides efficient retrieval and adaptive management, prioritizing important and safety-critical events. The Refinement Module intervenes when initial predictions lack clarity, leveraging context from LTM to enhance accuracy in ambiguous scenarios.

In short, the agent's workflow, depicted in Figure \ref{fig:architecture}, begins with the Perception Module analyzing inputs using context from STM and feeding data into STM. A clarity score, based on vagueness, discrepancy, and confidence, determines whether the prediction proceeds directly to LTM or requires refinement. If refinement is triggered, the Refinement Module reprocesses the input with memory-derived insights from LTM for safe and accurate prediction. This architecture is well-suited for construction activities as it incorporates various mechanisms to prioritize safety, which is essential in those environments.

\subsection{Perception Module}
The Perception Module serves as the initial processing unit of the locomotion prediction agent, responsible for interpreting multimodal inputs to generate a prediction of the user’s locomotion mode. It processes a spoken command (e.g., "I’m climbing down") and a sequence of nine field-of-view (FOV) frames captured by smart glasses over a 1.5-second period, with 0.25 seconds before and 1.25 seconds after the command issuance. These frames provide chronological visual context that, combined with the command, enables the module to infer the user’s intended activity in construction settings. The module also generates embeddings for both the command text and the FOV frames, which support subsequent long-term memory operations.

To enhance prediction accuracy, the Perception Module incorporates context from Short-Term Memory (STM). Recent events stored in STM, such as prior locomotion modes, environmental conditions, and objects or obstacles the user interacted with, inform the interpretation of the current command and frames for a more consistent and safer transition. For example, if the user was recently ascending a construction ladder, this context ensures consistency when interpreting a command like "I’m climbing down." Conversely, if the command is incorrectly given—such as saying "I’m going to walk" while at the top of a ladder—the module prioritizes a safer decision by relying on visual evidence and STM context. 

\begin{figure}
\centering
\includegraphics[width=.9\textwidth]{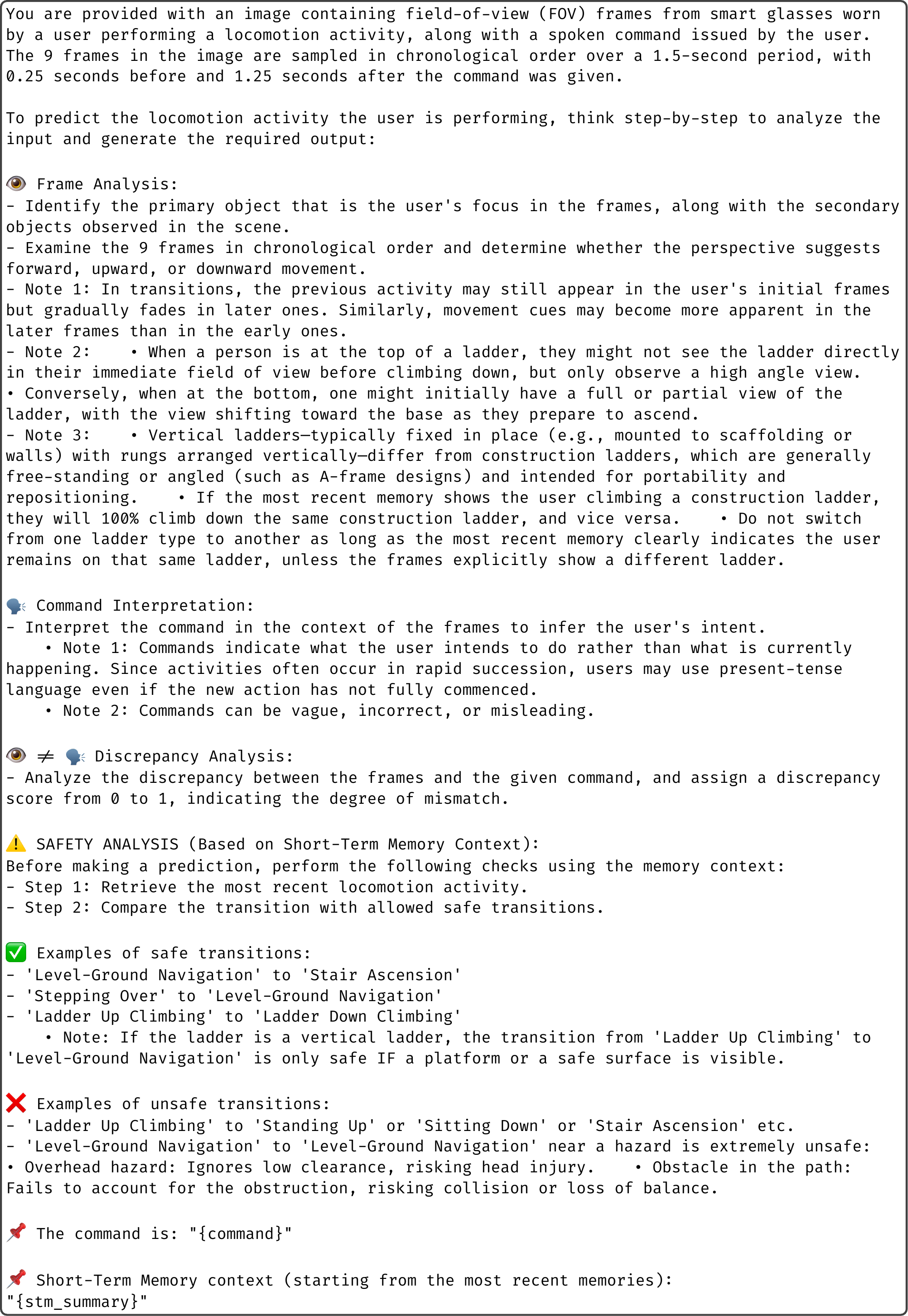}
\caption{Perception Prompt Used by the Perception Module}
\label{fig:perception_prompt}
\end{figure}

The module employs a detailed prompt, shown in Figure \ref{fig:perception_prompt}, to systematically guide the Large Language Model (LLM) in analyzing the inputs using Chain-of-Thought (CoT) reasoning. This structured approach instructs the LLM to inspect frames, interpret the command, and validate safety, producing a JSON-constrained response. The output includes a CoT reasoning trace, the predicted locomotion mode (e.g., "Construction Ladder Down Climbing"), and scores assessing vagueness (the command’s clarity), discrepancy (the mismatch between command and visual input), importance (the event’s criticality based on safety risks or relevance), and confidence (the LLM’s certainty in the prediction’s accuracy), along with scene context (environments such as indoor or outdoor), identified primary objects and obstacles, and a concise summary of the command and key visual details.

\subsection{Memory Systems}
\subsubsection{Short-Term Memory (STM)}

Short-Term Memory (STM) serves as a transient buffer, storing recent locomotion-related events to provide real-time context for both the Perception Module and the Refinement Module. It retains data within a configurable retention window, capturing details such as the user’s prior locomotion modes, environmental conditions, and interactions with objects or obstacles. This temporal scope ensures the agent maintains awareness of the user’s ongoing activity sequence, crucial for interpreting commands and validating transitions in dynamic settings.

STM operates as a time-sensitive record, continuously updated with each new perception output. To manage capacity and relevance, STM employs a pruning mechanism that removes entries beyond the retention window, ensuring only the most recent and pertinent information remains available. For example, if a user transitions from "Level-Ground Navigation" to "Ladder Up Climbing," STM retains this sequence, enabling the agent to detect inconsistencies or unsafe shifts, such as an abrupt command to "Sit Down" while atop a ladder.

The primary role of STM is to enhance prediction consistency and safety by supplying immediate context to the Perception and Refinement Modules. It enables cross-referencing of current inputs against recent events—such as confirming a ladder descent follows an ascent—or identifying discrepancies that may signal command errors. Particularly, STM underpins the safety analysis outlined in Figure \ref{fig:perception_prompt}, where its context is formatted as ``At \{timestamp\}: \{locomotion\_mode\} in a \{environment\} environment, interacting with a \{primary\_object\},'' derived from prior perception outputs. This structure enables checks like flagging unsafe shifts by retrieving the latest locomotion mode and comparing it to allowed transitions.

\subsubsection{Long-Term Memory (LTM)}
Long-Term Memory (LTM) acts as a comprehensive repository for all locomotion-related events, enabling the agent to draw on historical context for informed decision-making. Unlike Short-Term Memory (STM), which handles transient data, LTM maintains an enduring record of past events through a vector database. This database stores events as vector embeddings—semantic representations derived from textual and visual inputs—alongside associated metadata, with events categorized as either safety-critical (e.g., ladder transitions) or routine based on their locomotion mode. Safety-critical events are prioritized during processing to ensure their prominence in decision-making.

Retrieval from LTM employs a multi-vector search using cosine similarity, which measures the angular distance between the embedding of the current perception and those stored in memory to identify contextually relevant events. To reconcile potential mismatches between textual commands and visual inputs, the system dynamically weights text and image similarities using the discrepancy score (\(d\)). Text similarity is weighted by \(1 - d\), while image similarity is weighted by \(d\), allowing visual cues to take precedence when significant discrepancies arise, thus resolving ambiguities effectively.

Ranking of retrieved events relies on a composite score, calculated as \( \text{Composite Score} = (w_s \cdot \text{similarity}) + (w_i \cdot \text{importance}) + (w_c \cdot \text{confidence}) - (w_d \cdot \text{discrepancy} + w_v \cdot \text{vagueness}) \), where \( w_s \), \( w_i \), \( w_c \), \( w_d \), and \( w_v \) are configurable weights that adjust the influence of similarity (a weighted blend of text and image similarities), importance (a measure of an event’s criticality based on safety risks or relevance), and confidence (reflecting the LLM's prediction confidence), while applying a penalty for vagueness and discrepancy, reduced for safety-critical events, to ensure that events critical to safety, with lower vagueness and discrepancy, are ranked higher.

Memory management within LTM maintains efficiency and relevance through adaptive strategies. Each event’s importance score decays exponentially over time, with safety-critical events decaying more slowly to preserve their significance, while routine events fade more quickly. Periodic pruning eliminates routine events whose importance falls below a threshold, preventing memory overload while retaining safety-critical entries. Frequent retrieval of an event boosts its importance, enhancing its retention based on sustained utility.

\subsection{Refinement Module}
The Refinement Module enhances the locomotion prediction agent’s decision-making by re-evaluating ambiguous inputs. It activates when the initial prediction’s clarity score—computed as \( w_v \cdot (1 - \text{vagueness}) + w_d \cdot (1 - \text{discrepancy}) + w_c \cdot \text{confidence} \), where \( w_v \), \( w_d \), and \( w_c \) are configurable weights—falls below a dynamic threshold, indicating potential uncertainty or safety risks. This dynamic threshold increases over time as the agent’s memory systems accumulate more reliable events, ensuring that refinement is triggered only when necessary to avoid unnecessary computation for clear inputs.

The refinement process reanalyzes the spoken command and FOV frames, integrating them with context derived from both Short-Term Memory and Long-Term Memory. This enriched context is incorporated into a structured prompt for the Large Language Model, which extends the Perception Module’s prompt by including LTM-derived insights (provided as the retrieved locomotion mode along with a summary of the command and key visual details derived from the LLM output for each event), to generate a refined JSON-constrained response. This output includes an updated locomotion mode prediction, revised scores, and a reasoning trace. The module plays a critical role in maintaining safety and accuracy, particularly in construction environments where high vagueness or discrepancy in inputs could lead to unsafe predictions.

\section{Evaluation}
\subsection{Dataset}

\begin{figure}
    \centering
    \begin{subfigure}[b]{0.32\textwidth}  
        \includegraphics[width=\textwidth]{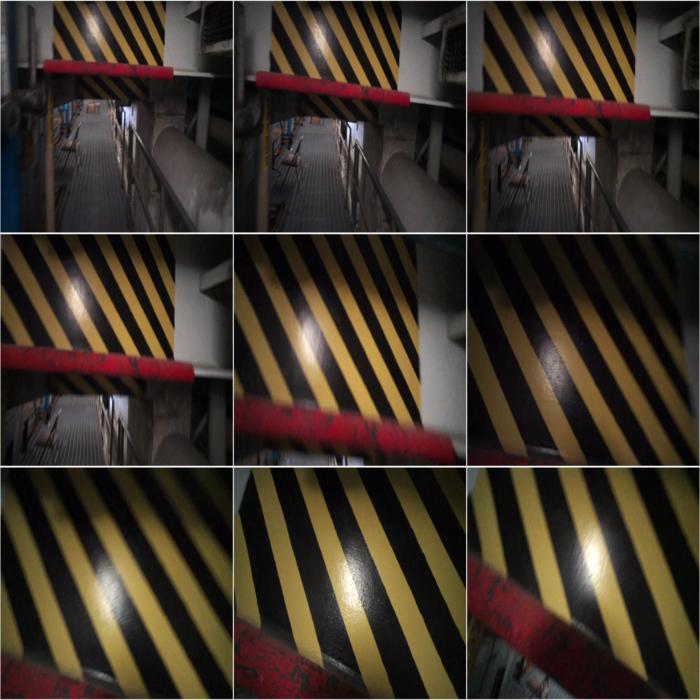}
        \caption{Low Space Navigation}
        \label{fig:low_space}
    \end{subfigure}
    \hfill
    \begin{subfigure}[b]{0.32\textwidth}  
        \includegraphics[width=\textwidth]{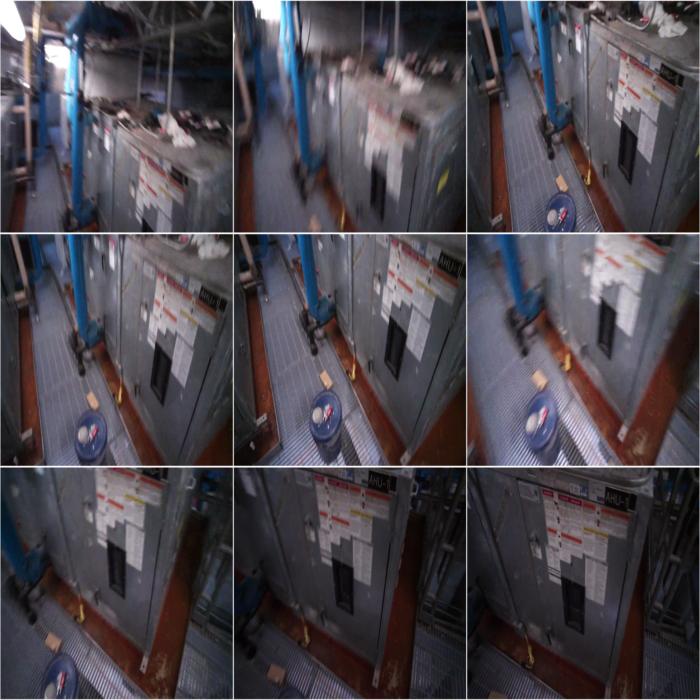}
        \caption{Sitting Down}
        \label{fig:sitting}
    \end{subfigure}
    \hfill
    \begin{subfigure}[b]{0.32\textwidth}  
        \includegraphics[width=\textwidth]{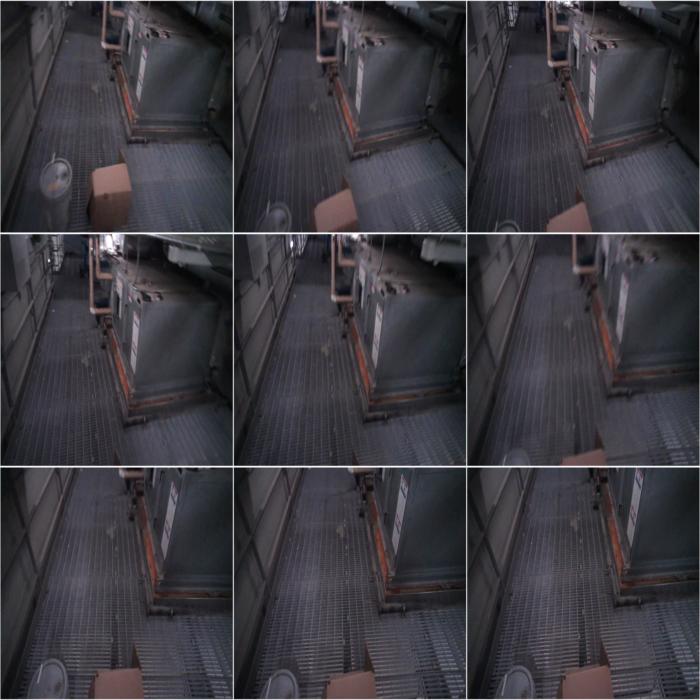}
        \caption{Stepping over Box}
        \label{fig:stepping_box}
    \end{subfigure}
    
    \vspace{0.2cm}

    \begin{subfigure}[b]{0.32\textwidth}  
        \includegraphics[width=\textwidth]{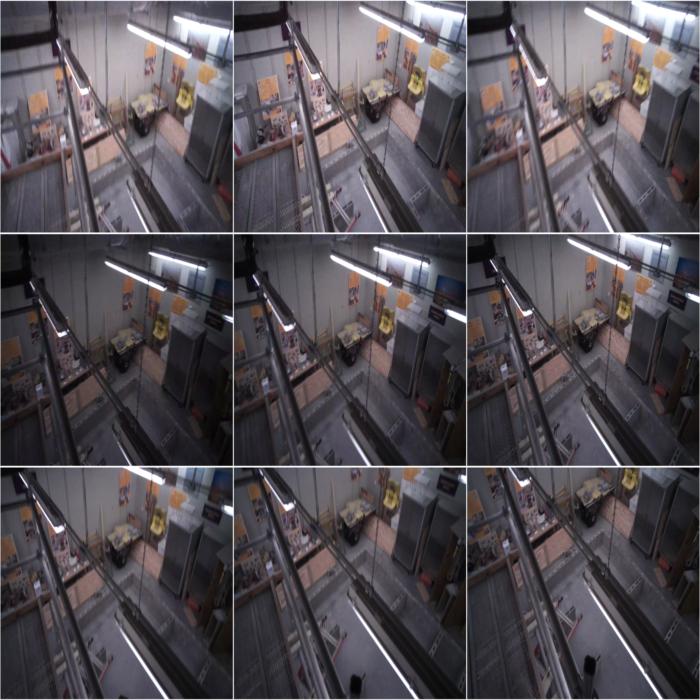}
        \caption{Vertical Ladder\\ Down Climbing}
        \label{fig:ladder_down}
    \end{subfigure}
    \hfill
    \begin{subfigure}[b]{0.32\textwidth}  
        \includegraphics[width=\textwidth]{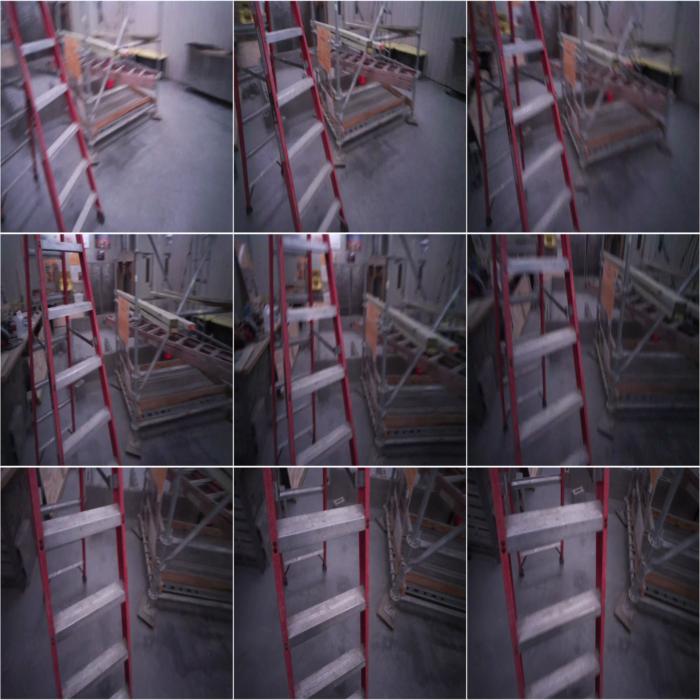}
        \caption{Construction Ladder\\ Up Climbing}
        \label{fig:const_up}
    \end{subfigure}
    \hfill
    \begin{subfigure}[b]{0.32\textwidth}  
        \includegraphics[width=\textwidth]{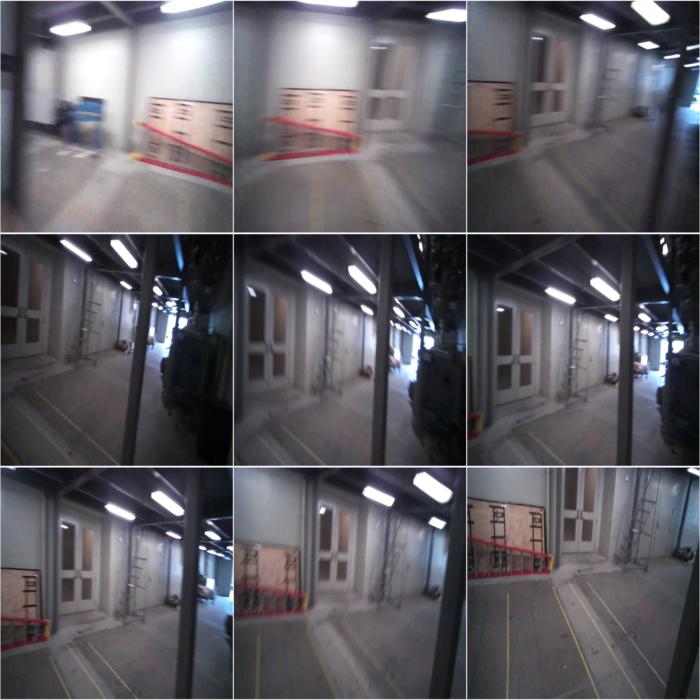}
        \caption{Level-ground \\Navigation}
        \label{fig:level_ground}
    \end{subfigure}

    \caption{Examples of FOV frames}
    \label{fig:locomotion_examples}
\end{figure}

We collected data within a environment tailored to simulate real-world construction scenarios, encompassing a range of locomotion modes related to construction workflows. The dataset includes field-of-view (FOV) frames processed as \(700 \times 700\) pixel grid images, each capturing nine sequential frames over a 1.5-second period—0.25 seconds before and 1.25 seconds after command issuance—as exemplified in Figure \ref{fig:locomotion_examples}. Table \ref{tab:locomotion_modes} lists the studied locomotion modes, which span vertical and construction ladder climbing, level-ground walking, stair ascension and descension, low-space navigation, obstacle navigation, sitting, and standing.

\begin{figure}
    \centering
    \includegraphics[width=\textwidth]{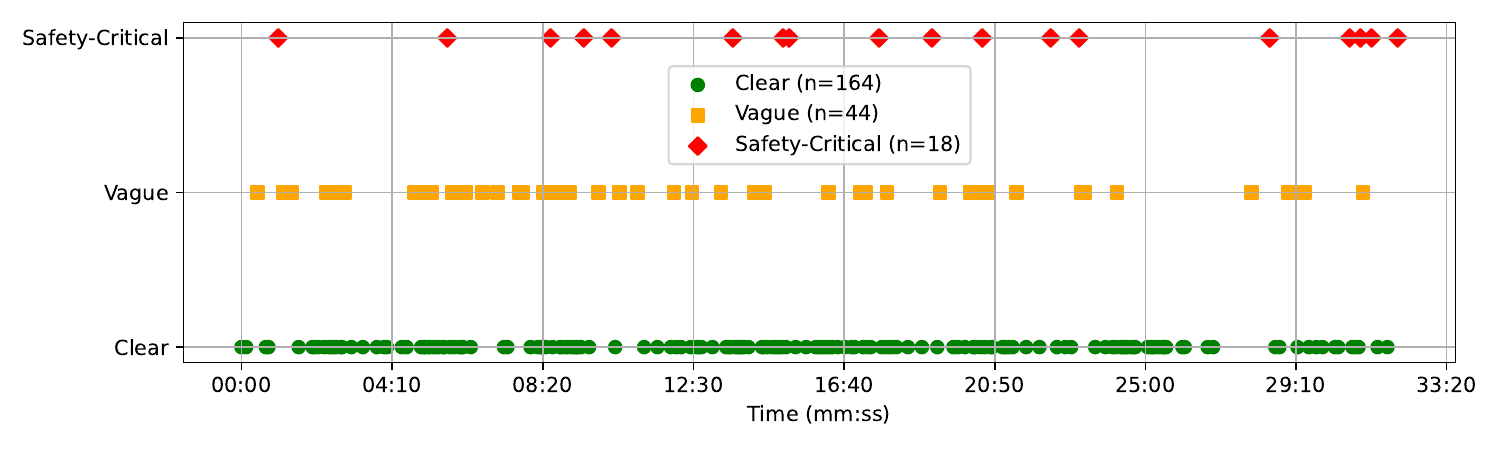}
    \caption{Distribution of Command Types in the Dataset}
    \label{fig:command_distribution}
\end{figure}

The dataset also incorporates three distinct sets of spoken commands to evaluate the system’s robustness amidst real-world complexities—clear, vague, and safety-critical—as detailed in Table \ref{tab:command_examples}. Clear commands provide precise instructions, such as ``I'm walking'' for Level-Ground Navigation, while vague commands like ``I'm heading up'' for Stair Ascension test the agent’s ability to handle ambiguity. Safety-critical commands, such as ``I'm walking forward'' when atop a construction ladder (intended for Construction Ladder Down Climbing), were included to evaluate performance in high-risk scenarios. Notably, we incorporated safety-critical commands for the most critical events—ladder climbing, obstacles, low space, and stairs—where incorrect commands could lead to major safety risks. Figure \ref{fig:command_distribution} depicts the distribution of these command types in the dataset.

\begin{table}
\centering
\caption{Studied Locomotion Modes}
\label{tab:locomotion_modes}
\begin{tabular}{l}
\toprule
Locomotion Mode \\
\midrule
Construction Ladder Down Climbing (CDwn)\\
Construction Ladder Up Climbing (CUp) \\
Vertical Ladder Down Climbing (VDwn) \\
Vertical Ladder Up Climbing (VUp) \\
Level-Ground Navigation (LGN) \\
Low Space Navigation (LSN) \\
Sitting Down (SD) \\
Standing Up (SU) \\
Stair Ascension (SAsc) \\
Stair Descension (SDsc) \\
Stepping over Box (SoB)\\
Stepping over Pipe (SoP) \\
\bottomrule
\end{tabular}
\end{table}

\begin{table}
\centering
\caption{Examples of Commands}
\label{tab:command_examples}
\begin{tabular}{lll}
\toprule
Command & Locomotion Mode & Command Type \\
\midrule
"I'm walking." & Level-Ground Navigation & Clear \\
"I want to climb this vertical ladder." & Vertical Ladder Up Climbing & Clear \\
"I'll step over this box." & Stepping over Box & Clear \\
"I'm moving." & Level-Ground Navigation & Vague \\
"I'm heading up." & Stair Ascension & Vague \\
"I'm walking over these." & Stepping over Pipe & Vague \\
"I'm going down." & Vertical Ladder Down Climbing & Vague \\
"I'm walking forward." & Construction Ladder Down Climbing & Safety-Critical \\
"I'm moving upright." & Low Space Navigation & Safety-Critical \\
"I'll keep walking." & Stepping over Box & Safety-Critical \\
"I'm going to lean forward." & Stair Descension & Safety-Critical \\
\bottomrule
\end{tabular}
\end{table}

\subsection{Configuration}

The locomotion prediction agent employs selected models and parameters to process multimodal inputs and manage memory, as outlined in Table \ref{tab:config_params}. The system utilizes \textit{gpt-4o} as the Large Language Model (LLM), with a maximum token limit of 600 and a temperature of 0.7 for perception tasks to balance exploration and precision in initial predictions. For refinement tasks, a lower temperature of 0.5 is applied to favor deterministic outputs, ensuring higher reliability when resolving ambiguities or safety-critical scenarios. Text embeddings are generated using the \textit{text-embedding-ada-002} model, while FOV frames are encoded with \textit{openai/clip-vit-large-patch14}, each producing embeddings of 768 dimensions processed separately for retrieval. Short-Term Memory (STM) retains events for 45 seconds, and Long-Term Memory (LTM) leverages ChromaDB as the vector database for storing and retrieving events, prioritizing similarity with a weight of 0.65, complemented by importance and confidence weights of 0.2 and 0.15, respectively, while applying penalty weights of 0.3 for discrepancy and 0.2 for vagueness, reduced by a factor of 0.7 for safety-critical events; LTM maintenance ensures relevance with decay rates of 0.005 for safety-critical events and 0.03 for routine events, alongside a pruning threshold of 0.1. The clarity score, guiding refinement activation, is computed with weights of 0.3 for vagueness, 0.5 for discrepancy, and 0.2 for confidence, with a threshold increasing from 0.35 to 0.75 over cycles.

\begin{table}
\centering
\caption{Main Configuration Parameters}
\label{tab:config_params}
\small
\begin{tabular}{ll}
\toprule
Parameter & Value \\
\midrule
LLM Model & \textit{gpt-4o} \\
Max Tokens & 600 \\
Perception Temperature & 0.7 \\
Refinement Temperature & 0.5 \\
Text Embedding Model & \textit{text-embedding-ada-002} \\
Image Embedding Model & \textit{openai/clip-vit-large-patch14} \\
Individual Embedding Dimension & 768 \\
STM Retention Threshold (s) & 45 \\
LTM Retrieval Top K & 5 \\
LTM Similarity Weight & 0.65 \\
LTM Importance Weight & 0.2 \\
LTM Confidence Weight & 0.15 \\
LTM Discrepancy Penalty Weight & 0.3 \\
LTM Vagueness Penalty Weight & 0.2 \\
LTM Safety-Critical Penalty Reduction & 0.7 \\
LTM Safety-Critical Decay Rate & 0.005 \\
LTM Routine Decay Rate & 0.03 \\
LTM Prune Importance Threshold & 0.1 \\
Clarity Vagueness Weight & 0.3 \\
Clarity Discrepancy Weight & 0.5 \\
Clarity Confidence Weight & 0.2 \\
Clarity Threshold Min & 0.35 \\
Clarity Threshold Max & 0.75 \\
Clarity Ramp per Cycle & 0.01 \\
\bottomrule
\end{tabular}
\end{table}

\subsection{Metrics}
We evaluated the locomotion prediction agent using precision, recall, and F1-score, weighted to account for class imbalance, to measure classification performance. Additionally, the Brier Score and Expected Calibration Error (ECE) were employed to assess prediction reliability and confidence alignment.

\subsection{Ablation Studies}
Ablation studies assessed the contributions of Short-Term Memory (STM) and Long-Term Memory (LTM) through three configurations: No Memory (NoMem), relying solely on the Perception Module without memory context; STM Only (STMOnly), incorporating recent context; and STM and LTM Combined (STM+LTM), the full system with recent and historical insights. Each configuration was tested across the dataset to evaluate precision, recall, and F1-score.

\section{Results and Discussion}

\subsection{Impact of Memory Systems}

The ablation study in Table \ref{tab:ablation_study_performance} compares the agent’s performance under three conditions: NoMem, STMOnly, and STM+LTM, each evaluated over 226 samples. The NoMem condition yields a weighted precision of 0.81, recall of 0.70, and F1-score of 0.73, indicating baseline performance without contextual memory. Adding STM improves these metrics to a weighted precision of 0.86, recall of 0.81, and F1-score of 0.81, while integrating both STM and LTM further elevates them to 0.92, 0.90, and 0.90, respectively. This progression demonstrates that STM provides effective immediate context to improve locomotion mode inference, and LTM adds historical relevance of similar events, resulting in a substantial F1-score increase from 0.73 to 0.90.

\begin{figure}
    \centering
    \includegraphics[width=0.85\textwidth]{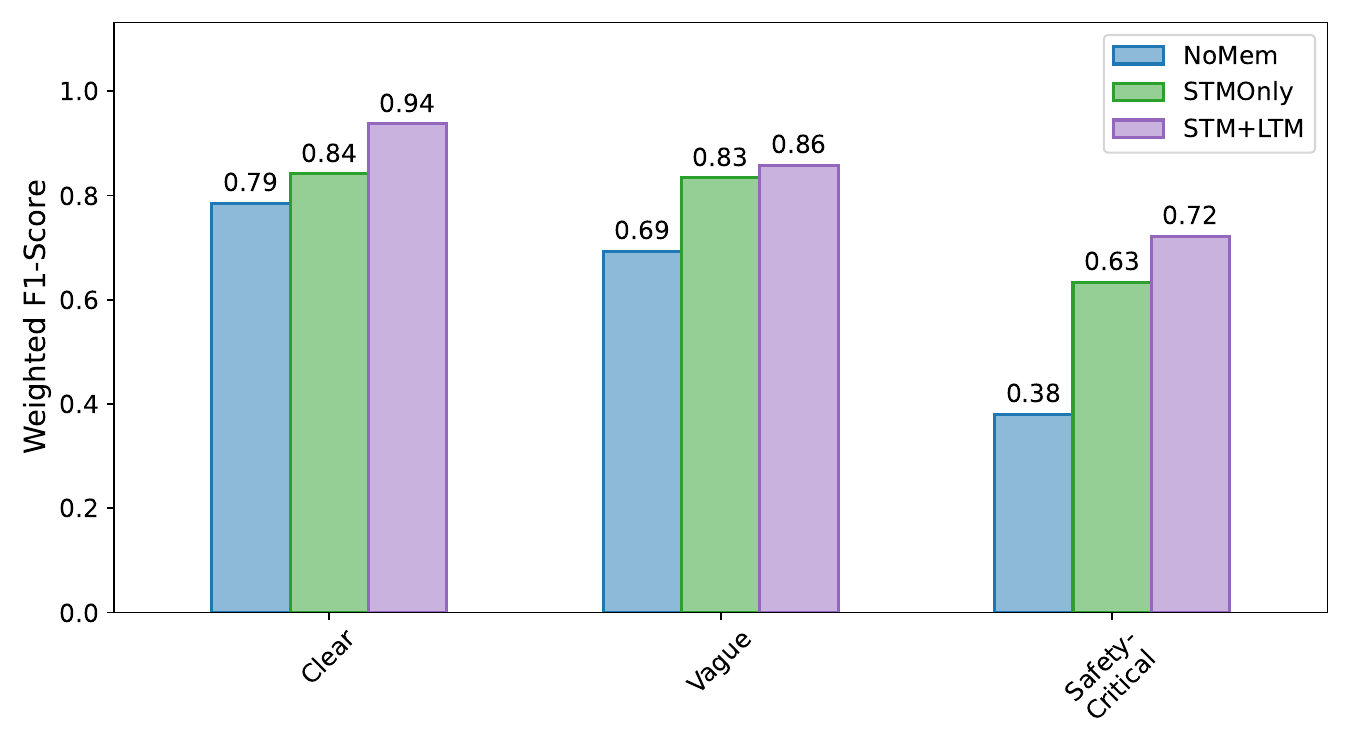}
    \caption{Weighted F1-Score by Command Type}
    \label{fig:command_type_f1}
\end{figure}

\begin{figure}
    \centering
    \includegraphics[width=0.75\textwidth]{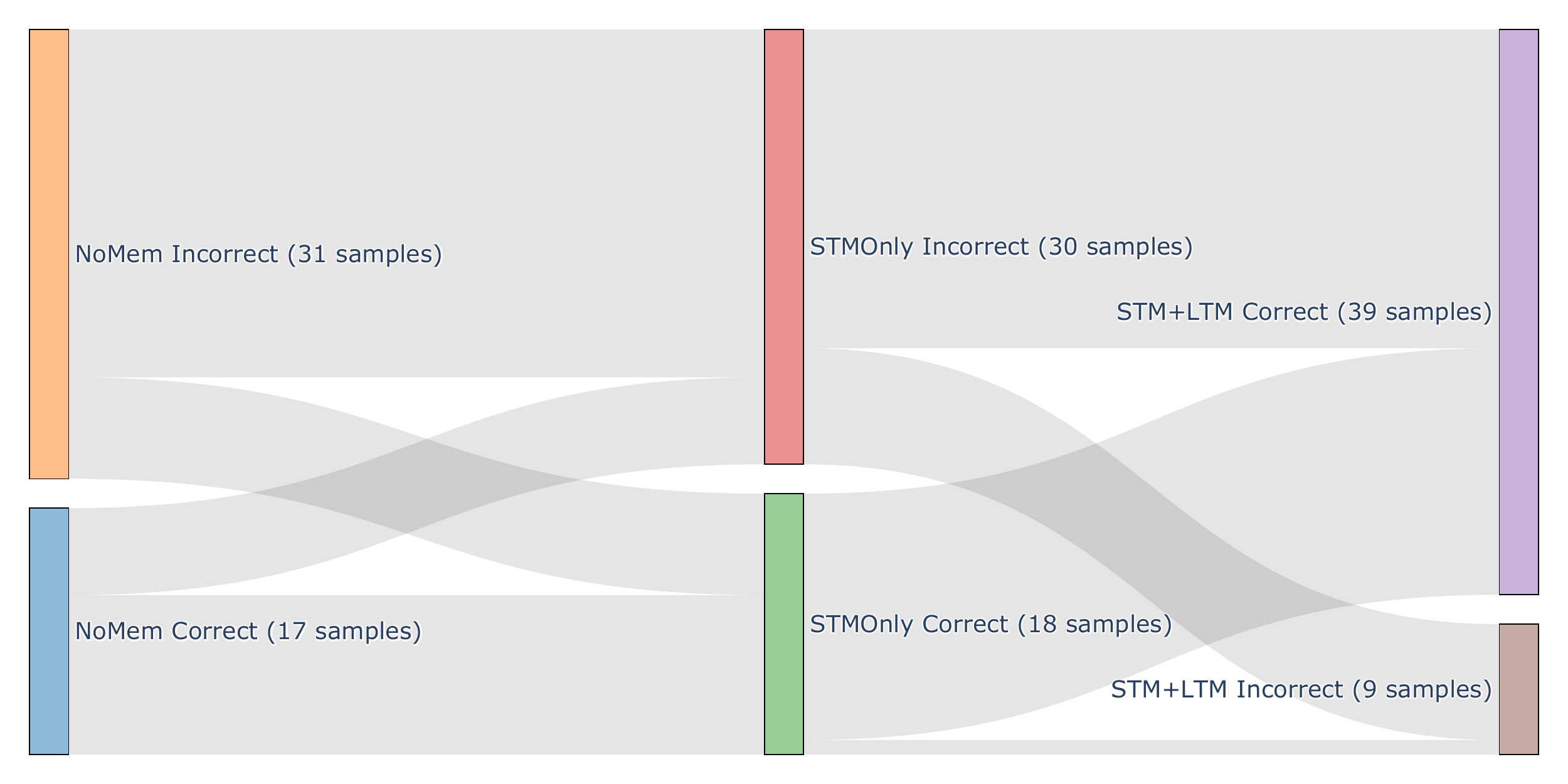}
    \caption{Prediction Outcome Change from NoMem to STMOnly to STM+LTM (Only for Samples Undergoing Refinement Using LTM)}   \label{fig:refinement_change}
\end{figure}

Figure \ref{fig:command_type_f1} breaks down this performance by command type—clear, vague, and safety-critical—highlighting memory’s differential impact. For clear commands, the weighted F1-score rises from 0.79 (NoMem) to 0.84 (STMOnly) and 0.94 (STM+LTM), showing near-perfect accuracy with full memory support. Vague commands improve from 0.69 to 0.83 and 0.86, with STM contributing the largest gain, resolving ambiguities by leveraging context from most recent previous activities. Safety-critical commands, starting at a low 0.38, reach 0.63 with STMOnly and 0.72 with STM+LTM, indicating significant improvement but persistent challenges in high-stakes scenarios. Figure \ref{fig:refinement_change} focuses on a subset of samples undergoing LTM refinement, revealing a shift from 17 correct and 31 incorrect predictions (NoMem) to 18 correct and 30 incorrect (STMOnly), and finally to 39 correct and 9 incorrect (STM+LTM). This highlights LTM’s refinement module as a key factor in correcting initial errors within this group, reducing incorrect predictions by over two-thirds.

\begin{table}
\centering
\caption{Ablation Study Performance Metrics (Weighted Average)}
\label{tab:ablation_study_performance}
\begin{tabular}{lcccc}
\toprule
Condition & Precision & Recall & F1-Score & Support \\
\midrule
NoMem & 0.81 & 0.70 & 0.73 & 226 \\
STMOnly & 0.86 & 0.81 & 0.81 & 226 \\
STM$+$LTM & \textbf{0.92} & \textbf{0.90} & \textbf{0.90} & 226 \\
\bottomrule
\end{tabular}
\end{table}

Table \ref{tab:calibration_metrics} further elucidates the impact of memory systems on prediction reliability by presenting calibration metrics across the three conditions. The Brier Score, measuring the mean squared difference between predicted probabilities and actual outcomes, decreases from 0.244 in the NoMem condition to 0.169 with STMOnly, and further to 0.090 with STM+LTM, indicating a progressive improvement in prediction calibration as memory systems are incorporated. Similarly, the Expected Calibration Error (ECE), which quantifies the misalignment between prediction confidence and accuracy, reduces from 0.222 to 0.133 and 0.044 across the same conditions, demonstrating that STM and LTM enhance the agent’s ability to align its confidence with actual performance.

\begin{table}
\centering
\caption{Calibration Metrics (Brier Score and ECE)}
\label{tab:calibration_metrics}
\begin{tabular}{lcc}
\toprule
Condition & Brier Score & ECE \\
\midrule
NoMem & 0.244 & 0.222 \\
STMOnly & 0.169 & 0.133 \\
STM+LTM & 0.090 & 0.044 \\
\bottomrule
\end{tabular}
\end{table}

\subsection{Class-Specific Performance and Error Analysis}

\begin{figure}
    \centering
    \includegraphics[width=.9\textwidth]{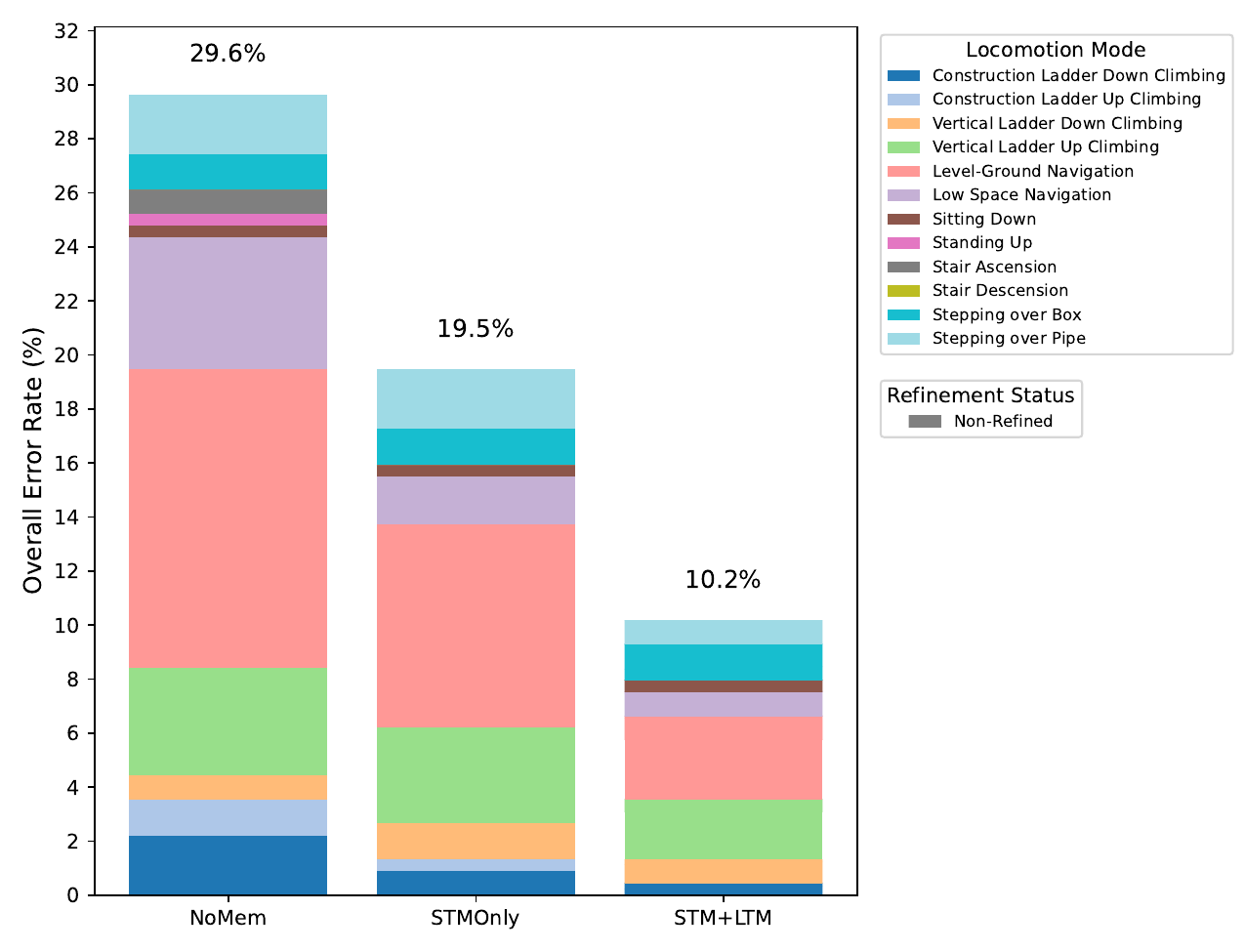}
    \caption{Overall Error Rate with Class-wise Breakdown}
    \label{fig:error_rate}
\end{figure}

Figure \ref{fig:error_rate} presents an overall error rate analysis across three conditions—NoMem, STMOnly, and STM+LTM—depicted as stacked bar charts totaling 29.6\%, 19.5\%, and 10.2\%, respectively. The NoMem condition exhibits the highest error rate, with dominant contributions from Level-Ground Navigation and Low Space Navigation, reflecting their prevalence in the dataset. STMOnly reduces the overall error to 19.5\%, a 10.1 percentage point decrease, with a noticeable reduction in these two modes. The improvement in Construction Ladder Up and Down Climbing is also evident, demonstrating STM’s effectiveness in facilitating expected transitions, such as Construction Ladder Up followed by Construction Ladder Down Climbing, by leveraging recent contextual events. However, Vertical Ladder Up and Down Climbing remain challenging, with no significant improvement occurring in these modes. Several instances of obstacles, such as Stepping over Box and Pipe, also persist unchanged in STMOnly. STM+LTM further lowers the error rate to 10.2\%, compressing many errors and achieving zero error rates in modes like Construction Ladder Up Climbing and Stair Ascension; however, a significant portion of these errors, such as obstacles like Stepping over Box and Pipe and Vertical Ladder Down Climbing, are non-refined (hatched segments), originating from the perception module’s difficulty in identifying highly vague or discrepant events, thus limiting LTM’s ability to refine them, as evidenced by subsequent confusion patterns.

Focusing on the STM+LTM condition, Table \ref{tab:stm_ltm_class_performance} details a weighted F1-score of 0.90, indicating strong overall accuracy in predicting locomotion modes. The highest F1-scores are recorded for Standing Up (1.00), Stair Ascension (1.00), Low Space Navigation (0.95), Level-Ground Navigation (0.94), Sitting Down (0.92), Stepping over Pipe (0.92), Stair Descension (0.88), Construction Ladder Up Climbing (0.88), and Stepping over Box (0.86). In contrast, ladder-related activities exhibit lower performance. Construction Ladder Down Climbing scores 0.80 with a precision of 0.75 and recall of 0.86, performing better than the vertical ladder classes but still below the overall average. Vertical Ladder Down Climbing records an F1-score of 0.62, with a precision of 0.50 and recall of 0.83, while Vertical Ladder Up Climbing has an F1-score of 0.67, with a precision of 1.00 but a recall of 0.50, indicating that all predictions are correct yet many instances are missed. The support column reveals a class imbalance, with Level-Ground Navigation comprising 100 samples, while other classes include a lower number of samples. This distribution aligns with the continuous nature of the dataset, where Level-Ground Navigation serves as a frequent transitional state between other tasks. It is also worth mentioning that safety-critical scenarios usually occur for ladder activities, especially during down climbing, where the agent should provide adjusted locomotion modes when commands can lead to risky consequences.

\begin{figure}
    \centering
    \includegraphics[width=\textwidth]{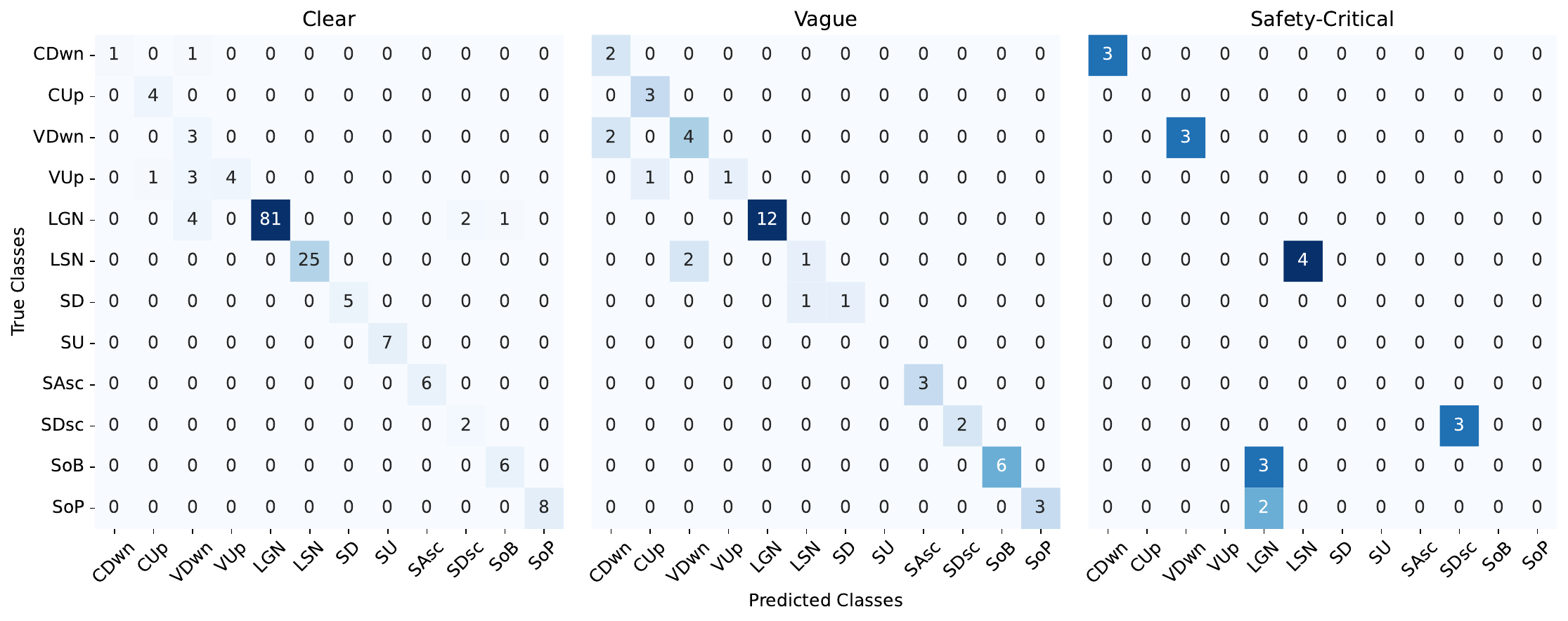}
    \caption{Confusion Matrices of STM+LTM Predictions Across Clear, Vague, and Safety-Critical Commands (darker colors indicate higher prediction counts)}
    \label{fig:cm_stm_ltm}
\end{figure}

Figure \ref{fig:cm_stm_ltm} elucidates these performance disparities through confusion matrices across command types, focusing on misclassifications where errors occur. For clear commands, Construction Ladder Down Climbing is misclassified, with 1 out of 2 instances predicted as Vertical Ladder Down Climbing, while Vertical Ladder Up Climbing shows significant confusion, with 4 out of 8 instances correctly predicted, but 1 misclassified as Construction Ladder Up Climbing and 3 as Vertical Ladder Down Climbing. Level-Ground Navigation, with 81 out of 88 correct, experiences errors where 4 instances are misclassified as Vertical Ladder Down Climbing, 2 as Stair Descension, and 1 as Stepping over Box. For vague commands, Vertical Ladder Up Climbing (1 out of 2 correct) is misclassified as Construction Ladder Up Climbing (1 instance), Low Space Navigation (1 out of 3 correct) has 2 instances misclassified as Vertical Ladder Down Climbing, and Sitting Down (1 out of 2 correct) has 1 instance misclassified as Low Space Navigation. In safety-critical commands, Stepping over Box (0 out of 3 correct) is entirely misclassified as Level-Ground Navigation, and Stepping over Pipe (0 out of 2 correct) is similarly misclassified as Level-Ground Navigation, while Construction Ladder Down Climbing and Vertical Ladder Down Climbing remain fully accurate (3 out of 3 each). These misclassification patterns correlate with the higher non-refined error rates and lower F1-scores of ladder and obstacle classes, indicating their susceptibility to confusion with Level-Ground Navigation, particularly under vague or safety-critical conditions.

\begin{table}
\centering
\caption{Performance Metrics of STM+LTM Across Classes}
\label{tab:stm_ltm_class_performance}
\begin{tabular}{lcccc}
\toprule
Class & Precision & Recall & F1-Score & Support \\
\midrule
Construction Ladder Down Climbing & 0.75 & 0.86 & 0.80 & 7 \\
Construction Ladder Up Climbing & 0.78 & 1.00 & 0.88 & 7 \\
Vertical Ladder Down Climbing & 0.50 & 0.83 & 0.62 & 12 \\
Vertical Ladder Up Climbing & 1.00 & 0.50 & 0.67 & 10 \\
Level-Ground Navigation & 0.95 & 0.93 & 0.94 & 100 \\
Low Space Navigation & 0.97 & 0.94 & 0.95 & 32 \\
Sitting Down & 1.00 & 0.86 & 0.92 & 7 \\
Standing Up & 1.00 & 1.00 & 1.00 & 7 \\
Stair Ascension & 1.00 & 1.00 & 1.00 & 9 \\
Stair Descension & 0.78 & 1.00 & 0.88 & 7 \\
Stepping over Box & 0.92 & 0.80 & 0.86 & 15 \\
Stepping over Pipe & 1.00 & 0.85 & 0.92 & 13 \\
\midrule
Overall (Weighted Avg) & 0.92 & 0.90 & 0.90 & 226 \\
\bottomrule
\end{tabular}
\end{table}

\subsection{Perception Score Distributions and Their Role in LTM Decision-Making}

\begin{figure}
    \centering
    \includegraphics[width=\textwidth]{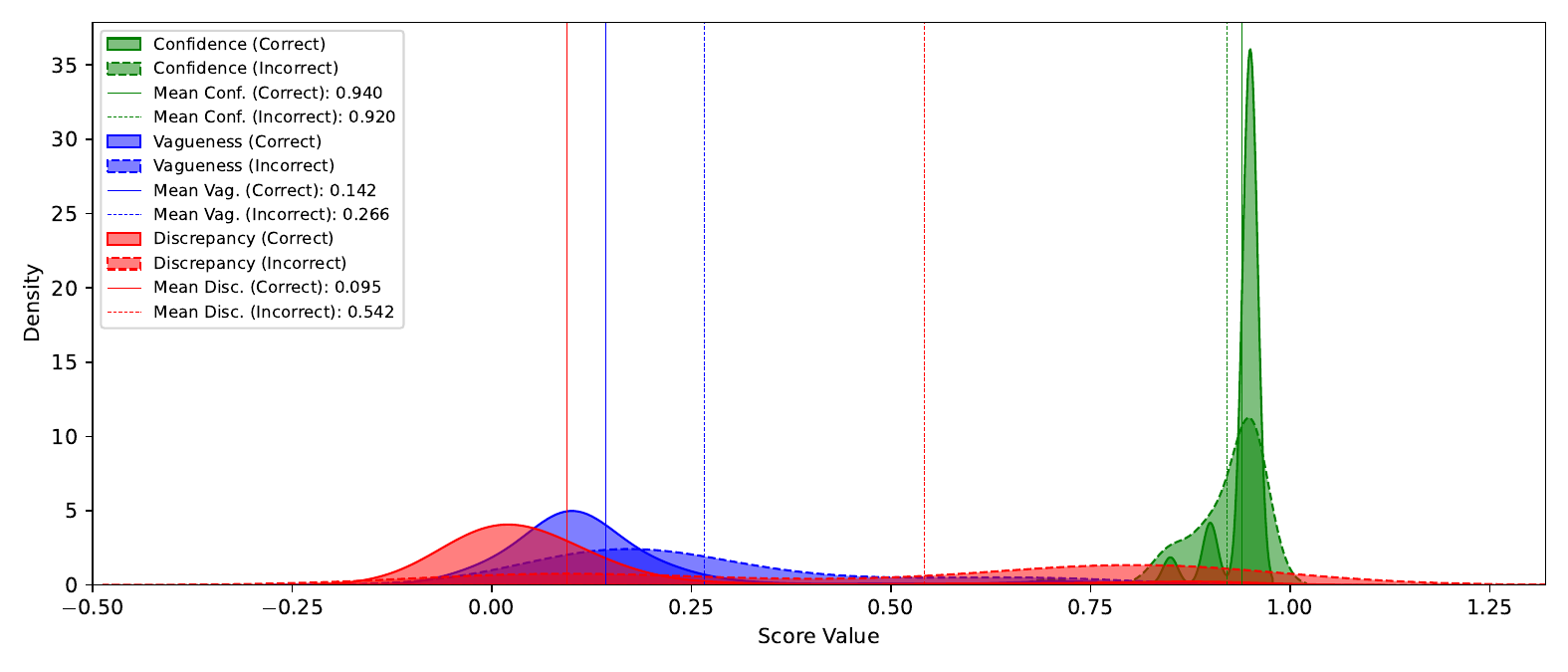}
    \caption{Kernel Density Estimation of Perception Module Scores (confidence, discrepancy, and vagueness) Stratified by Prediction Correctness}
    \label{fig:score_distributions}
\end{figure}

Figure \ref{fig:score_distributions} illustrates kernel density estimation (KDE) plots of the perception module’s confidence, vagueness, and discrepancy scores, stratified by prediction correctness (correct vs. incorrect), to assess their influence on long-term memory (LTM) decision-making. The confidence distribution for correct predictions peaks near the mean of 0.940, with a narrow, high-density profile near 1.0, while incorrect predictions peak near the mean of 0.920, showing a slightly broader tail extending below 0.9, indicating limited separation between the two. In contrast, the vagueness distribution for correct predictions peaks near the mean of 0.142, with a tight cluster near 0.0, whereas incorrect predictions peak near the mean of 0.266, with a wider spread toward higher values, suggesting a notable shift. The discrepancy distribution exhibits the most pronounced difference, with correct predictions peaking near the mean of 0.095 and tightly clustered near 0.0, while incorrect predictions peak near the mean of 0.542, with a significant rightward shift and extended tail, indicating strong differentiation.

These distributional trends underscore the discriminative power of the perception scores, justifying their weighted roles in LTM decision-making. Discrepancy demonstrates the highest discriminative power, with a mean difference of 0.447 between incorrect and correct predictions, supporting its dominant weight of 0.5 in the clarity score formula (\(w_d = 0.5, w_v = 0.3, w_c = 0.2\)), which determines whether events are transferred directly to LTM or subjected to refinement. This weight aligns with discrepancy’s ability to identify significant command-visual mismatches, triggering refinement for more ambiguous cases. Vagueness, with a mean difference of 0.124, exhibits moderate discriminative power, warranting its intermediate weight of 0.3, as it effectively flags events with higher uncertainty for LTM processing. Confidence, with a minimal mean difference of 0.020, shows weak discriminative power, justifying its lower weight of 0.2, reflecting its limited utility in distinguishing correct from incorrect predictions. Furthermore, discrepancy’s strong separation supports its use in LTM retrieval, where it dynamically adjusts embedding weights (\(w_{\text{text}} = 1 - \text{discrepancy}, w_{\text{image}} = \text{discrepancy}\)), prioritizing image-based similarity for events with high discrepancy. The penalty term in the LTM composite score, incorporating discrepancy and vagueness with weights of 0.3 and 0.2 respectively, leverages their discriminative trends to rank retrieved events, ensuring priority for less ambiguous instances, consistent with the agent’s design objectives.

\section{Conclusion}

This paper presents a novel locomotion mode prediction agent aimed at enhancing exoskeleton assistance for construction workers by leveraging Large Language Models (LLMs) augmented with memory systems. The proposed system tackles the critical challenge of intended locomotion recognition in the dynamic, safety-critical construction environment, using multimodal inputs—spoken commands and visual data from smart glasses—to accurately predict locomotion modes. The agent’s architecture, comprising a Perception Module, Short-Term Memory (STM), Long-Term Memory (LTM), and a Refinement Module, collectively addresses the unpredictability and variability of construction tasks.

Evaluation results highlight the efficacy of this approach, with memory system integration markedly enhancing prediction performance. The ablation study shows the baseline configuration without memory achieving a weighted F1-score of 0.73, rising to 0.81 with STM, and reaching 0.90 with both STM and LTM combined. This gain is especially critical for vague and safety-critical commands, where STM resolves ambiguities with immediate context, and LTM refines predictions using historical patterns. Calibration metrics reinforce this improvement: the Brier Score falls from 0.244 to 0.090, and the Expected Calibration Error drops from 0.222 to 0.044, reflecting not only higher accuracy but also better-aligned confidence. The Perception Module’s score distributions, particularly the highly discriminative discrepancy scores, validate the system’s design by guiding refinement decisions effectively.

Despite these advancements, challenges persist, particularly with locomotion modes like vertical ladder climbing and obstacle navigation, where misclassifications remain notable in safety-critical scenarios. These limitations underscore the need to enhance the Perception Module’s ability to produce more reliable scores, ensuring LTM refinement is triggered effectively when ambiguity arises. Additionally, deploying the system in real-time construction settings with user feedback would yield valuable insights into its practical efficacy and usability.

In conclusion, this research presents a significant advancement in locomotion mode prediction for construction-related activities, offering a robust framework for safer and more effective high-level human-exoskeleton collaboration. By harnessing LLMs and memory systems, the agent addresses the unique demands of construction environments and enhances worker safety. Beyond construction, the principles and architecture developed here hold promise for broader applications in industries requiring adaptive, context-aware assistive systems, paving the way for future innovations in human-robot interaction.

\section{Acknowledgments}
This material is based upon work supported by the National Science Foundation under Grant No. 2222881. Any opinions, findings, and conclusions or recommendations expressed in this material are those of the authors and do not necessarily reflect the views of the National Science Foundation.

\bibliographystyle{plainnat}
\bibliography{main}
\end{document}